\documentclass[10pt,twocolumn,letterpaper]{article}

\usepackage[pagenumbers]{cvpr} 

\usepackage[dvipsnames]{xcolor}

\definecolor{cvprblue}{rgb}{0.21,0.49,0.74}
\usepackage[pagebackref,breaklinks,colorlinks,citecolor=cvprblue]{hyperref}
\usepackage{booktabs} 
\usepackage{amsmath}
\usepackage{colortbl}
\usepackage{amsthm}
\usepackage{dsfont}
\usepackage{indentfirst}
\usepackage{gensymb}
\usepackage[accsupp]{axessibility}

\def\MethodName{\textcolor{black}{InNeRF360}}
\title{\MethodName{}: Text-Guided 3D-Consistent Object Inpainting\\on 360\degree~Neural Radiance Fields}

\author{Dongqing Wang \quad Tong Zhang\thanks{Corresponding author.} \quad Alaa Abboud \quad Sabine Süsstrunk\\
School of Computer and Communication Sciences, EPFL\\
Lausanne, Switzerland\\
}

\def\MethodName{\textcolor{black}{InNeRF360}}
\newcommand{\norm}[1]{\left\lVert#1\right\rVert}

\newenvironment{tight_itemize}{
\begin{itemize}[leftmargin=15pt]
  \setlength{\topsep}{0pt}
  \setlength{\itemsep}{0pt}
  \setlength{\parskip}{0pt}
  \setlength{\parsep}{0pt}
}{\end{itemize}}

\begin{document}
\maketitle

\begin{abstract}
We propose \MethodName{}, an automatic system that accurately removes text-specified objects from 360\degree Neural Radiance Fields (NeRF). 
The challenge is to effectively remove objects while inpainting perceptually consistent content for the missing regions, which is particularly demanding for existing NeRF models due to their implicit volumetric representation. 
Moreover, unbounded scenes are more prone to floater artifacts in the inpainted region than frontal-facing scenes, as the change of object appearance and background across views is more sensitive to inaccurate segmentations and inconsistent inpainting. 
With a trained NeRF and a text description, our method efficiently removes specified objects and inpaints visually consistent content without artifacts.
We apply depth-space warping to enforce consistency across multiview text-encoded segmentations, and then refine the inpainted NeRF model using perceptual priors and 3D diffusion-based geometric priors to ensure visual plausibility.  
Through extensive experiments in segmentation and inpainting on 360\degree and frontal-facing NeRFs, we show that our approach is effective and enhances NeRF's editability. Project page: \url{https://ivrl.github.io/InNeRF360/}. 
\end{abstract}

\vspace{-0.3cm}
\section{Introduction}
\label{sec:intro}
\begin{figure}
  \centering
  \includegraphics[width=0.8\linewidth]{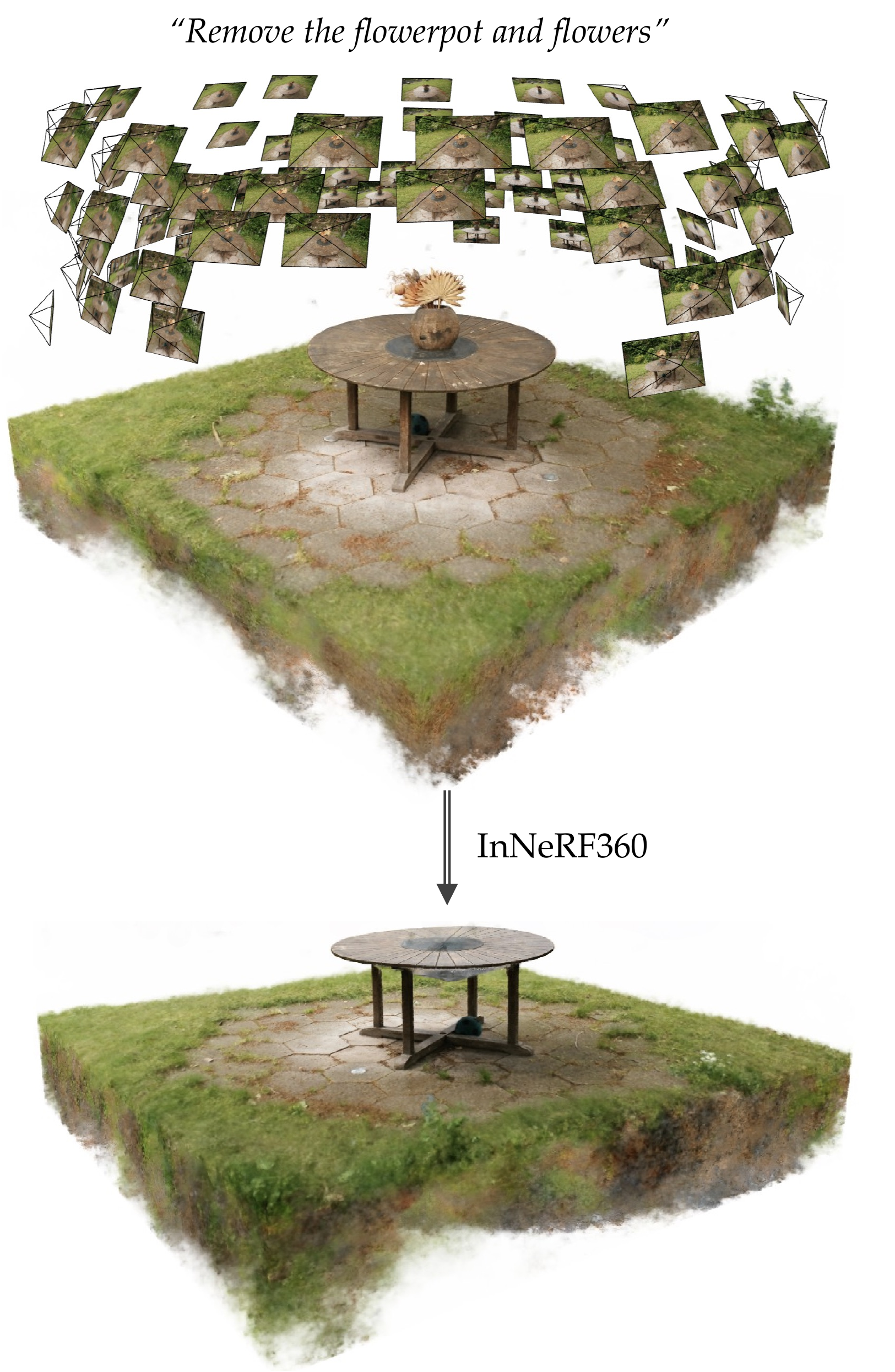}
  \caption{Given a pre-trained NeRF and a text to remove specific objects (e.g.\textit{``Remove the flowerpot and flowers''}), \MethodName{} produces accurate multiview object segmentations, and outputs an inpainted NeRF with visually consistent content. }        
  \vspace{-0.6cm}
  \label{fig:front}
\end{figure}

Recreating and manipulating real-world scenarios is one of the main focuses of Virtual and Augmented Reality (VR/AR) applications. 
Neural Radiance Field (NeRF) and its variants~\cite{mildenhall2020nerf, barron2022mip, muller2022instant} can efficiently model 360\degree real-world scenes for photorealistic novel view synthesis. Consequently, they have the potential to become widely accessible tools for representing the 3D world. 

A desired feature of such applications is the ability to modify the content of the created scene, including object removal. 
However, direct inpainting in the NeRF framework is intractable due to the implicit representation of the captured scenes encoded through the weights of multilayer perceptrons (MLP), which hinders explicit user control of scene contents.
For explicit NeRF variants, the scene representation of radiance fields contains ambiguous surface that is hard to segment with a bounding region.efore, obtaining an accurate segmentation for objects within is nontrivial.

Existing works on selecting~\cite{ren-cvpr2022-nvos} or removing~\cite{spinnerf, liu2022nerf, weder2022removing} objects in a trained NeRF tackle the 3D problem by utilizing 2D input. 
These methods begin with sparse user inputs, then use 2D image/video segmentation~\cite{matterport_maskrcnn_2017, cheng2021stcn} for multiview segmentation and inpaint on RGB-D sequence.
They are restricted to frontal-facing viewing angles, as the input scribbles or masks cannot extrapolate across different viewpoints in 360\degree~scenes where the shape of the object can change drastically.  
Moreover, in the case of object occlusions on 360\degree~scenes, inpainting 2D depth maps is inadequate for geometric supervision as it leads to inconsistencies in scene geometry.

In contrast, \MethodName{} is an inpainting pipeline with a depth-guided segmentation method dedicated to accurate object-level editing on 360\degree~scenes. Instead of extrapolating sparse 2D input to 3D, 
\MethodName{} encodes text input into a promptable segmentation model, Segment Anything Model (SAM)~\cite{kirillov2023segment}, leveraging its accurate semantic segmentation.
Our method is driven by the intuition that an object's semantic identity is \textit{more consistent}  over different viewpoints than its geometry. 
However, text-based 2D semantic segmentations may not always maintain consistency across views. To overcome this, we refine object masks using inverse depth space warping in 3D space across viewpoints, utilizing the consistent 3D positioning of objects.

Using view-consistent masks, we train a NeRF from scratch on the multiview inpainting from a 2D image inpainter~\cite{rombach2022high}. 
The multiview training images slightly differ in the inpainted regions, accumulating into cloudy artifacts, i.e. floaters~\cite{Nerfbusters2023}, in the new NeRF.
To eliminate floaters, we finetune the scene guided by 3D diffusion priors trained on extensive geometric shapes~\cite{chang2015shapenet} to determine whether density should be removed or incremented in a local voxel. 
The texture for the removed region is optimized by contextual appearance priors~\cite{zhang2018unreasonable} from the surrounding regions of the segmentation. 
This creates a perceptually consistent 3D inpainted region that seamlessly blends into the scene.
Extensive experiments show that \MethodName{} can effectively inpaint both 360\degree~\cite{barron2022mip} and front-facing~\cite{mildenhall2019local} real-world scenes, with the potential to be extended into 3D editing.

To summarize, our contributions are as follows:
\begin{tight_itemize}
    \item \MethodName{} is the first work to achieve text-guided object inpainting in 360\degree~NeRF scenes, ensuring visually consistent inpainted regions.
    \item Our approach efficiently generates multiview consistent 2D segmentation for 3D object inpainting through depth-warping refinement on initialized masks.
    \item We incorporate a 3D diffusion network as local geometric priors to remove artifacts in the inpainted region.
    
\end{tight_itemize}

\section{Related Works}
\noindent\textbf{Image Inpainting.}
Recent advancements in image inpainting focus on filling in masked regions to create visually consistent images. 
This primarily involves generative models like Generative Adversarial Networks (GANs)~\cite{pathak2016context, zhao2021large, iizuka2017globally} and denoising diffusion models (DDPMs)~\cite{saharia2022palette, rombach2022high, nichol2021glide}. 
These models generate photorealistic predictions for the missing pixels. 
However, when applied to inpainting multiview renderings of 3D scenes~\cite{spinnerf, mirzaei2023reference}, they often generate different inpainted regions for nearby views.  
This is due to their lack of 3D understanding, presenting a challenge for inpainting 2D observations of 3D scenes. 
\MethodName{} leverages image inpainting to address inpainting 360\degree~scenes, ensuring view consistency across viewpoints.

\begin{figure*}
  \centering
  \includegraphics[width=0.9\linewidth]{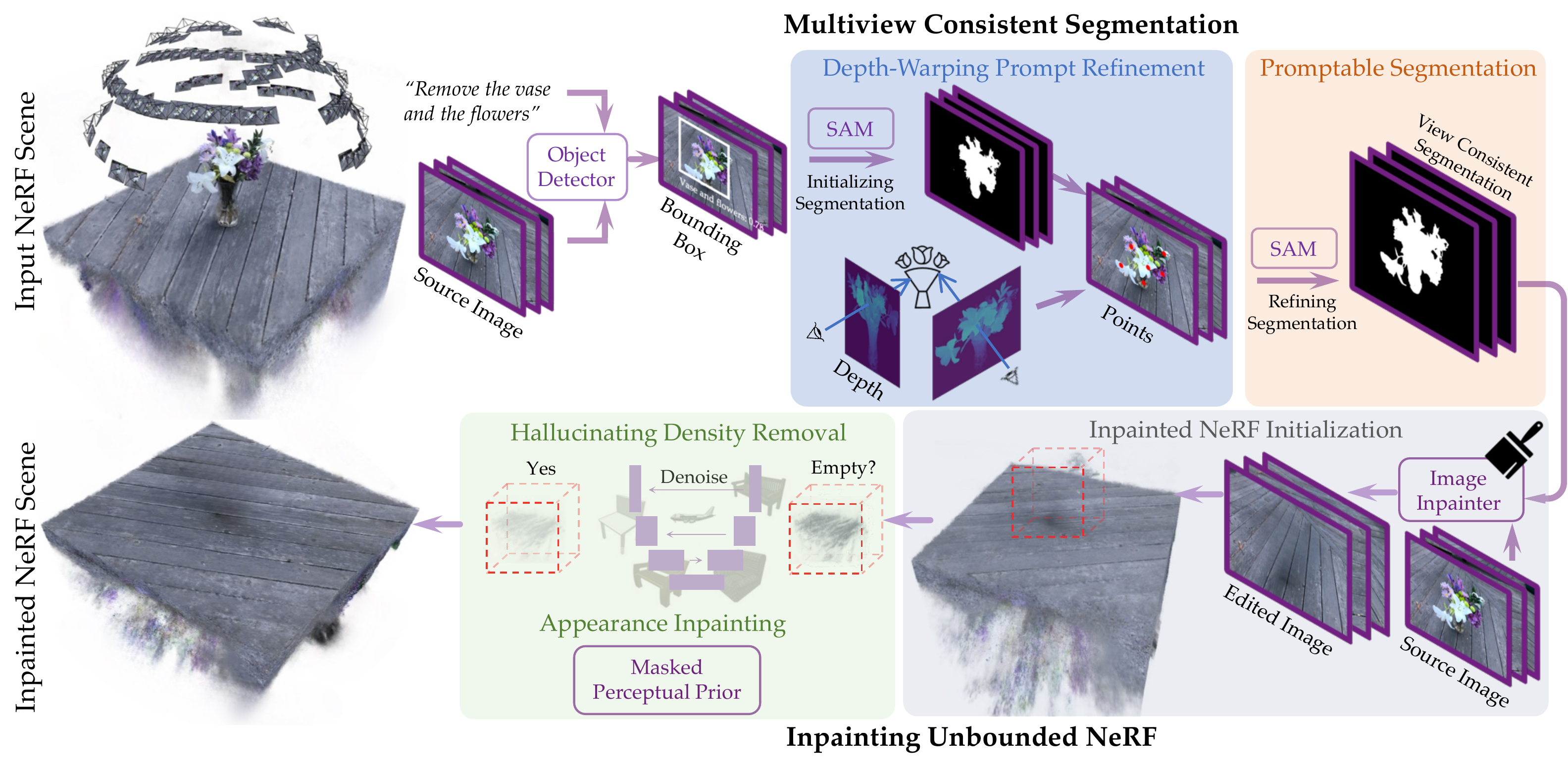}
  \vspace{-0.3cm}
  \caption{\textbf{Overview of \MethodName{} framework.} 1. \textit{Multiview Consistent Segmentation.} We initialize masks using bounding boxes from the object detector, which encodes both the source image and text. With rendered depth from the input NeRF, we apply depth-warping prompt refinement to iteratively update points for the Segment Anything Model (SAM) to output view-consistent 2D segmentations. 2. \textit{Inpainting 360\degree~NeRF.} We obtain edited images through image inpainter with the masks and source images to retrain the inpainted NeRF. We then finetune the new NeRF model using a geometric prior trained from a 3D diffusion model and a masked perceptual prior.}
  \vspace{-0.4cm}
  \label{fig:modelarch}
\end{figure*}

\noindent\textbf{Inpainting Neural Radiance Fields.}
Using neural radiance fields~\cite{mildenhall2020nerf} to represent 3D scenes has achieved high-quality, photo-realistic novel view synthesis. 
NeRFacto~\cite{tancik2023nerfstudio} is an architecture designed to optimize NeRF's performance on real-world data. 
It incorporates various recent advances in NeRF, including hash encoding~\cite{muller2022instant} and per-image appearance encoding~\cite{martin2021nerf}, among others. 

However, object removal presents a challenge in NeRF due to the implicit scene representation by the underlying neural networks.
Previous works~\cite{wang2022clip, mirzaei2022laterf} utilize the supervision of Contrastive Text-Image Pre-Training (CLIP)~\cite{radford2021learning}. They focus on inpainting a single object and cannot generalize to real-world scenes.
Methods utilizing depth-based approaches~\cite{liu2022nerf, weder2022removing, spinnerf, mirzaei2023reference} remove objects with user-drawn masks and depth sequences, and inpaint missing regions naively with 2D image inpainter on the training images of NeRF. 
These approaches are limited to front-facing scenes for two reasons.
First, their segmentation relies on the quality of initialize masks by supervised video object segmentation methods~\cite{zhang2022dino, oquab2023dinov2, caron2021emerging, cheng2021stcn}, which struggle on temporal consistency across frames for challenging cases such as transparent objects.
These methods cannot output accurate object masks in 360\degree~scenes, where object shapes change drastically across views.
Secondly, the chosen 2D inpainting methods output different inpainting across views. Such inconsistency results in floaters in the trained NeRF, and is much more pronounced on 360\degree~scenes than on frontal-facing ones. 
In contrast, \MethodName{} enhances the consistency of multiview segmentation by utilizing semantical identity and object 3D location consistency, thereby producing accurate masks for desired objects. 
Moreover, \MethodName{} is designed for 360\degree~NeRF inpainting by removing floaters (\cref{fig:float}) from the inpainted NeRF with geometric priors, and inpaint with contextual perception guidance.  

\noindent\textbf{Text-Guided 3D Editing.}
Given the popularity of text-conditioned image generative models, many works focus on generating 3D content with text instructions. Some rely on joint embeddings of CLIP to synthesize 3D meshes~\cite{michel2022text2mesh, mohammad2022clip} or neural radiance fields~\cite{jain2022zero, lee2022understanding}. Others distill a pre-trained diffusion model to optimize NeRF scenes in the latent space~\cite{poole2022dreamfusion, metzer2022latent}. 
These methods all suffer from having to map the inconsistent 2D diffusion model outputs to a 3D-consistent scene.
Instruct-NeRF2NeRF~\cite{haque2023instruct} edits renderings of a pre-trained NeRF model to preserve 3D consistency. 
However, it cannot remove scene objects or perform object-level editing as it operates in latent space for image editing. 
Our \MethodName{} operates in image space to accurately pinpoint and crop objects and allows removing an arbitrary number of objects from the scene through text instructions, while using 3D diffusion priors for local geometry finetuning to avoid the global inconsistency prior works exhibit.

\vspace{-0.2cm}
\section{Method}
\MethodName{} takes as input a trained NeRF with source images which the model trained on, and an instructive text. 
It outputs the inpainted 3D scene with the desired object(s) removed and filled with a visually consistent background without artifacts.  
Our pipeline is shown in~\cref{fig:modelarch}.

\subsection{Background: Neural Radiance Fields}
A Neural Radiance Field (NeRF) encodes a 3D scene as a function $f_{\theta}$ parametrized by an MLP with learnable parameters $\theta$, which maps a 3D viewing position $\mathbf{x}$ and its 2D direction $\mathbf{d}$ to the density $\sigma$ and a viewing-dependent color $\mathbf{c}$: $f_\theta: (\mathbf{x}, \mathbf{d}) \rightarrow (\sigma, \mathbf{c})$. Rendering a NeRF from a posed camera is done by sampling batches of rays for the camera pose, and rendering corresponding pixel colors for each ray. 
For each ray $\mathbf{r} = (\mathbf{o}, \mathbf{d})$, we sample an array of 3D points ($\mathbf{x}_i$, $t_i$), $i = 1, 2, \cdots, K$, where $ \mathbf{x}_i \in \mathbb{R}^2$ and $t_i$ is the depth. We query the MLP with these points along the ray for $\{\sigma_i\}^K_{i=1}$  and $\{\mathbf{c}_i\}^K_{i=1}$. 

The estimated RGB of ray $\hat{\mathbf{C}}(\mathbf{r})$ is obtained by alpha compositing~\cite{mildenhall2020nerf} the densities and colors along the ray: \vspace{-0.2cm}
\begin{equation}
    \hat{\mathbf{C}}(\mathbf{r})=\sum_{i=1}^{K}\alpha_i T_i\mathbf{c}_i,
    \label{eq:NeRF}
\end{equation}
where $T_i=1 - \exp(-\sigma_i\norm{t_i-t_{i-1}})$ is the ray transmittance between $\mathbf{x}_i$ and $\mathbf{x}_{i+1}$, and $\alpha_i=\prod_{j=1}^{i-1}T_i$ is the attenuation from ray origin to $\mathbf{x}_{i}$. The MLP is optimized through pixel loss for the distance between the estimated pixel value and the ground truth color.

\subsection{Multiview Consistent Segmentation}
The first stage is to obtain refined multiview segmentation masks for the objects 
to remove/edit given by the text input.
We take as input a pre-trained NeRF model and $N$ source images given by the set of $N$ source camera poses.

\begin{figure}[h]
  \centering
  \includegraphics[width=0.9\linewidth]{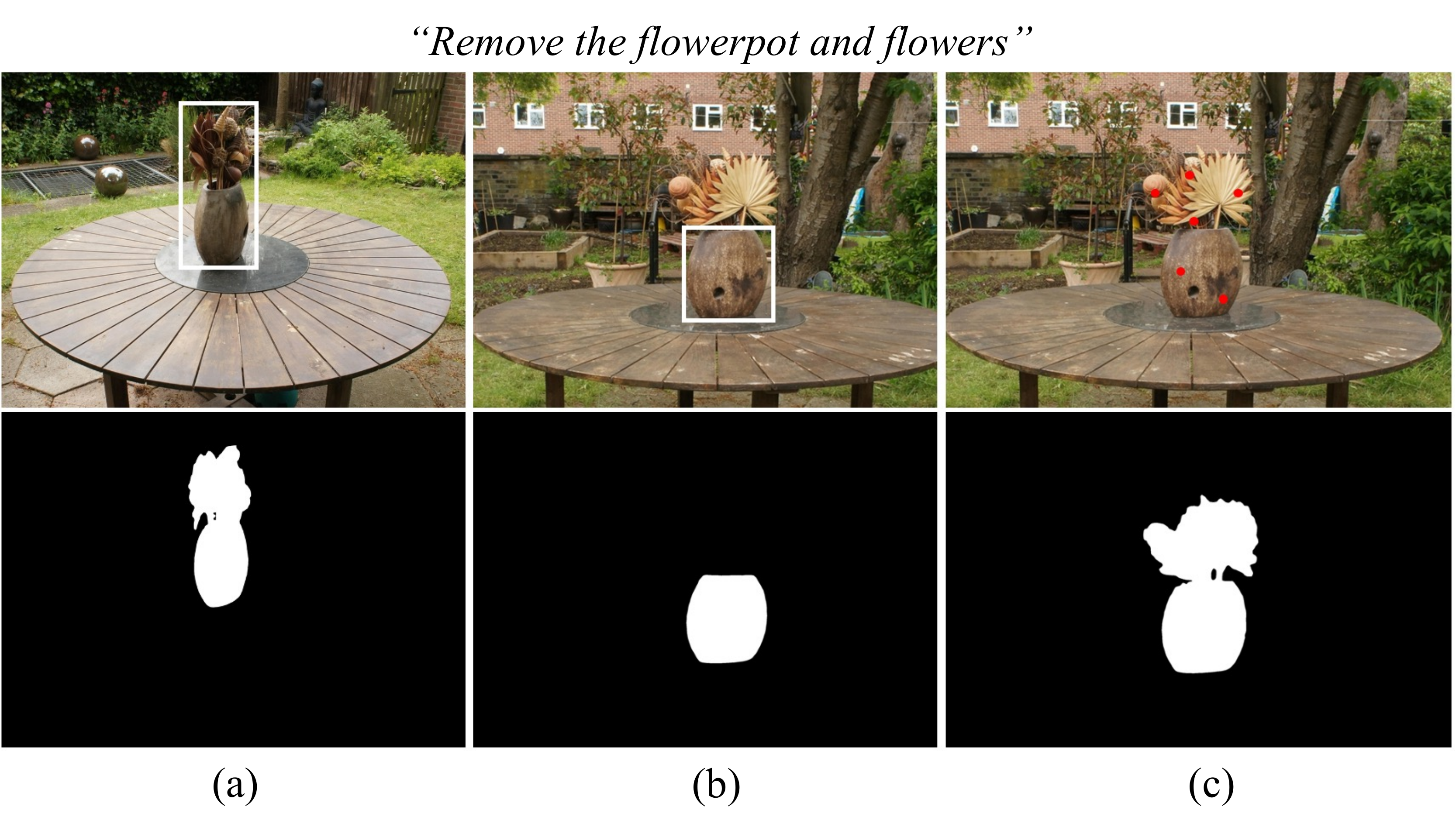}
  \vspace{-0.5cm}
  \caption{\textbf{Inconsistent bounding box across different views.} (a) and (b) are from the same dataset under the same instruction. However, the generated bounding boxes are different. After applying depth warping refinement (point prompts as red dots), (c) generates accurate segmentation. }
  \vspace{-0.4cm}
  \label{fig:depth}
\end{figure}

We initialize segmentations through the Segment-Anything Model (SAM)~\cite{kirillov2023segment} with the bounding boxes given by Grounded Language-Image Pre-training (GLIP) from a large-scale dataset of image-text pairs~\cite{li2022grounded}.
GLIP: $\mathcal{D}(\mathbf{I}, \mathbf{s}) = \{(B_q, p_q)\}^Q_{q=1}$ takes in an RGB image $\mathbf{I}\in \mathbb{R}^{H\times W\times3}$ and a text $\mathbf{s}$, and encodes each through respective image and language encoders. $Q$ is the number of bounding boxes. 
We then take these bounding boxes $\{B_q\}^Q_{q=1}$ from the model output in which $B_q = (l_q, r_q, u_q, d_q) \in\mathbb{R}^4$ with corresponding probability $p_q$. 
These boxes are sometimes inaccurate and fail to enclose the desired region, see~\cref{fig:depth} (b). Hence, we propose our depth-based prompt refinement for consistent segmentation.

\noindent\textbf{Depth-Warping Prompt Refinement.}
\label{sec:depth}
``Depth Warp'' operates on the sampled points to enhance segmentation accuracy. 
It leverages the depth information inherited from the input NeRF and projects these points back into the pixel space to align them with other 2D observations within the scene.
\textit{We thereby establish a cohesive depth constraint across different views.}
The depth for a sampled ray $\mathbf{r} = \mathbf{o} + t\mathbf{d}$ with origin $\mathbf{o}$ and direction $\mathbf{d}$ in a trained NeRF scene can be estimated via a modification of Eq.~\eqref{eq:NeRF}:  
\vspace{-0.3cm}
\begin{equation}
    D(\mathbf{r})=\sum_{i=1}^{I}\alpha_i T_i\cdot t_i,
    \label{eq:depth}
\end{equation} 
\vspace{-0.3cm}

For each training view, we randomly select $m$ other training views. 
From each selected view, we sample a fixed number $p$ of rays that correspond to pixels within the mask region. 
Given a ray $\mathbf{r}$ and its estimated depth from \cref{eq:depth}, we compute the 3D location of the selected point prompt. 
This information is then mapped back to the current training view. 
Rays and their corresponding point prompts whose depth is above a certain threshold are discarded, as they may represent background objects misclassified as foreground sections to be removed.
 If the corresponding pixel on the current view falls outside the masked region in the current view, we add this point prompt to the current view. 
 Subsequently, we pass this view with the new sets of point prompts to SAM for refined segmentation. 

By generating these ``out-of-the-box" point-based prompts for the segmentation model with our NeRF-based depth prior, we ensure that the object is accurately segmented even in cases where the initial bounding box information is insufficient or incomplete from certain viewpoints. This approach greatly increases the segmentation masks' accuracy for desired objects in the scene, see \cref{fig:depth}.

\vspace{-0.3cm}

\noindent\textbf{Promptable Segmentation}
\label{sec:mask}
After obtaining the refined point-based prompts, we employ SAM to identify and segment all the rendered observations $\{\mathbf{I}_n\}^N_{n=1}$ for refined masks $\{\mathbf{M}_n\}^N_{n=1}$. 
Specifically, SAM takes as input an image $\mathbf{I}$ and a prompt in the form of $j$ points pinpointing the object ${o_q}$, and produces an accurate segmentation mask $\mathbf{M_q}$ of the same size as $\mathbf{I}$: $\mathbf{M}_q = \texttt{SAM}(\mathbf{I}, o_q).$
For each image $\mathbf{I}_i \in \{\mathbf{I}_n\}^N_{n=1}$, we get a union binary mask for all the objects to be removed: 
\vspace{-0.3cm}
\begin{equation}
\mathbf{M}_i = \bigcup\limits_{q=1}^{Q}\mathbf{M}_{q}.
\end{equation}

\subsection{Inpainting 360\degree~NeRF}
\label{sec:nerf}
\begin{figure}[h]
  \centering
  \includegraphics[width=\linewidth]{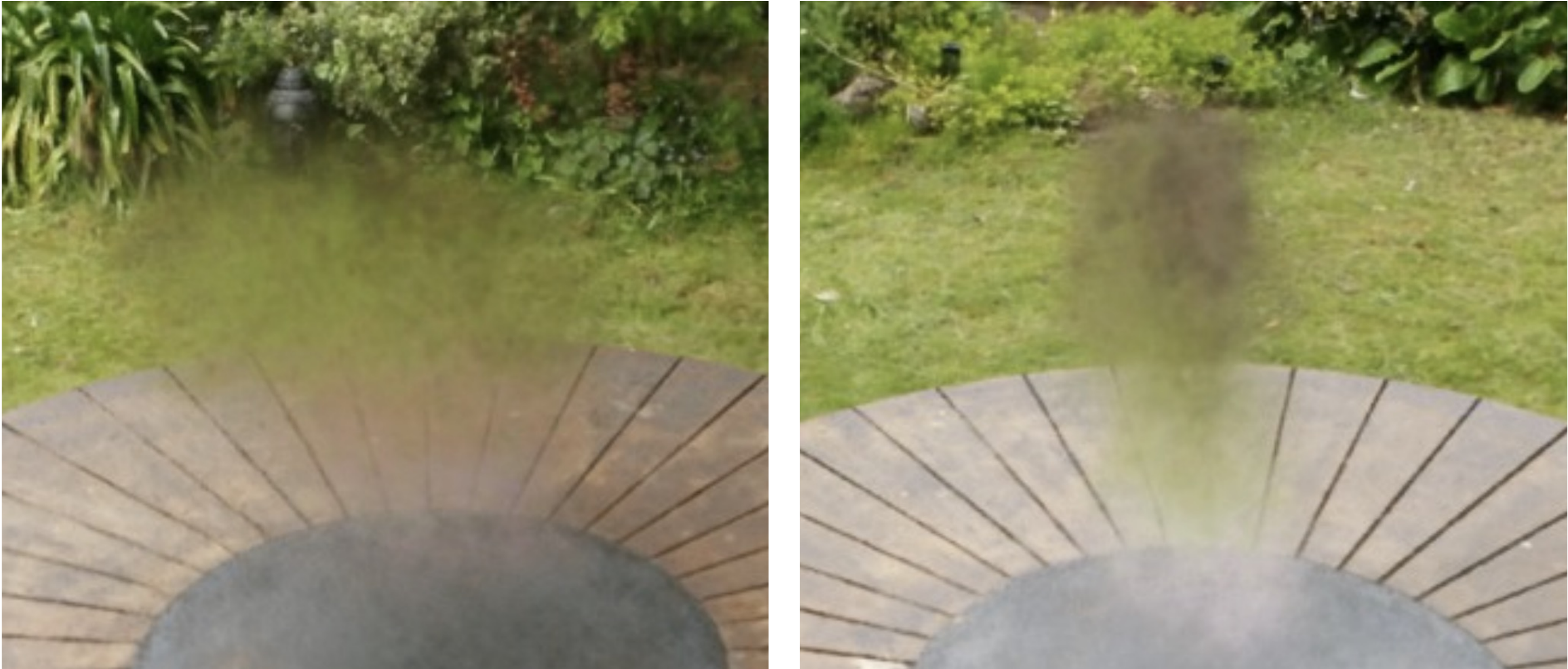}
  \vspace{-0.5cm}
  \caption{\textbf{Examples of artifacts in the initialized NeRF.} 2D inpaintings contain inconsistent inpainted pixels that accumulate in the 3D inpainted region and appear as floater artifacts. }
  \vspace{-0.4cm}
  \label{fig:float}
\end{figure}

\noindent\textbf{Inpainted NeRF Initialization.}
With the rendered observations $\{\mathbf{I}_n\}^N_{n=1}$ and corresponding masks $\{\mathbf{M}_n\}^N_{n=1}$, we adopt a 2D image inpainter~\cite{rombach2022high} to edit each observation as priors for the optimization. 
We then initialize the inpainted scene with Nerfacto~\cite{tancik2023nerfstudio}. This architecture is designed to optimize the performance of NeRF on real-world image captures.
However, each inpainted image has varying pixel-level content despite being perceptually plausible as a standalone image. Therefore, relying solely on RGB supervision leads to floaters in the NeRF~(\cref{fig:float}). To produce clean and perceptually consistent inpainting, we finetune the initialized NeRF with both geometry and appearance priors.

\medskip
\noindent\textbf{Hallucinating Density Removal.}
\label{sec:diffusion}
The density artifacts produced through inconsistent 2D inpainting can be seen as floaters. We train a denoising diffusion probabilistic model (DDPM)~\cite{NEURIPS2020_4c5bcfec} on ShapeNet~\cite{chang2015shapenet} to iteratively denoise a $m^3$ resolution voxel grid of discretized binary occupancy $x$ as a local 3D geometry prior: given a NeRF density $\sigma$ at timestep $t$, $x_t = 1$ if $\sigma > \rho$ else $x_t = -1$, where $\rho$ is a chosen threshold for whether a voxel is empty. For training, from each ShapeNet mesh we randomly select $N$ cubes that encompass 3\% to 8\% of the object bounding volume, and voxelize them to $m^3$ resolution. 
The loss function for the diffusion model is given by the MSE loss between the true noise $\epsilon$ and the predicted noise $\epsilon_{\theta}$ where $\theta$ parameterizes the diffusion model U-Net:
\vspace{-0.2cm} 
\begin{equation}
\mathcal{L}_{\text{ddpm}}(\theta) = \mathbb{E}_{t, x_0, \sigma}[\norm{\epsilon - \epsilon_\theta(\sqrt{\bar{\alpha}_{t}} x_0 + \sqrt{1- \bar{\alpha}_{t}}\epsilon, t)}_2^2],
\end{equation} 
where $t\in[1, 1000]$ is the number of timesteps in the noising diffusion process, and $\bar{\alpha}_{t} = \prod_{s=1}^t \alpha_{s}$ where $\alpha_{t}$ is a function of the timestep $t$ that parameterized transitions from ${x}_0$ to ${x}_{1000}$ (i.e. a noise schedule).

The Density Score Distillation Sampling (DSDS)~\cite{Nerfbusters2023} loss is defined to penalize regions with density $\sigma > \rho $ that the trained diffusion model deems as empty, and regions that are empty where $x_t = 1$ predicted by the model: \vspace{-0.2cm}
\begin{equation}
\mathcal{L}_{\text{DSDS}} = \sum_i u_i\sigma_i + (1-u_i)\max{(w-\sigma_i,0)},\vspace{-0.2cm}
\end{equation} 
where $u=\mathds{1}\{x_0 < 0\}$, and $w$ is a chosen upper limit for density $\sigma$ in occupied voxels.

The original sampling of the DSDS loss is by sampling a low-resolution density grid stored through ray bundles. 
During training, the grid selects the center of 3D cubes to be voxelized and enter the diffusion process. 
The grid gets updated by a visibility field determining what it sees in the NeRF scene. 
Our aim is to focus on removing floaters within the inpainted region. 
Therefore, we apply the refined segmentation mask from the image space to limit the visibility field to only look at the inpainted regions, and eliminate sampled rays corresponding to pixels outside the mask:\vspace{-0.3cm}
\begin{equation}
\mathcal{L}_{\text{geom}} = \sum_j (u_j\sigma_j + (1-u_j)\max{(w-\sigma_j,0)})\cdot V_j, \vspace{-0.2cm}
\label{eq:lossgeom}
\end{equation} 
where $V$ is an indicator function whose value is 1 for each voxel cube whose center is located within the current ``visible region'' given by the segmentation masks.
Consequently, we sample cubes near the inpainted region. The geometric prior trained on extensive shapes in~\cite{chang2015shapenet} preserves the original surface (e.g., the table supporting the removed flowerpot) while penalizing the floaters in the inpainted region created through inconsistent 2D inpainting. 
 
\noindent\textbf{Perceptually-Consistent Appearance Inpainting.}
\label{sec:perceptual}
The above method removes floaters created by inconsistent 2D inpainting, however, it does not produce visually consistent textures to fill in the removed region. Therefore, we utilize a patch-based loss~\cite{zhang2018unreasonable} as an appearance prior to enhancing the model's robustness and alleviate blurring effects.

Specifically, we sample all the pixels from the input inpainted images to get $W$ image patches $\{P_w\}^W_{w=1}$ with the size of $\upsilon^2$. These patches, $\{P_w\}^W_{w=1}$, can be divided into two non-overlapping groups of patches $P_{w_i}$ and $P_{w_o}$ depending on whether a patch \textit{contains} pixels within the inpainted region. For patches in $P_{w_o}$, we apply pixel-wise L1. This pixel loss is obtained by comparing the RGB values of each pixel $p$ in the inpainted patch, denoted as $\Tilde{\mathbf{C}}_p$, with the corresponding rendered RGB value, denoted as $\hat{\mathbf{C}}_p$,
\vspace{-0.3cm}
\begin{equation}
\mathcal{L}_{\textrm{pix}} = \frac{1}{\upsilon^2|P_{w_o}|}\sum_{p_\mathbf{r}\in P_{w_o}} \norm{\hat{\mathbf{C}}_p - \Tilde{\mathbf{C}}_p} _1.
\label{eq:img_loss}
\end{equation}

For patches in $P_{w_i}$ containing the inpainted region, we compute the perceptual similarity using LPIPS between the inpainted image patch $\Tilde{P}_I$ and the corresponding patch $P$ on the rendered image. We denote $\mathbf{C}_P$ as the set of pixel values in patch $P$, and define $\mathcal{L}_{\textrm{in}}$ as the inpainting loss.
\vspace{-0.3cm}
\begin{equation}
\mathcal{L}_{\textrm{in}} = \frac{1}{|P_{w_i}|}\sum_{P\in P_{w_i}} \text{LPIPS}(\mathbf{C}_P, \mathbf{C}_{\Tilde{P}_I}).
\label{eq:inpaint_loss}
\end{equation}

The inpainting loss measures the perceptual difference between the inpainted patch and the target inpainted image, while the pixel loss quantifies the pixel-level discrepancy between the inpainted and rendered RGB values. Together, these losses provide a comprehensive assessment of the reconstruction quality, accounting for both perceptual similarity and pixel-wise accuracy. 

\noindent\textbf{Loss Functions.}
Our optimization is the weighted sum of the geometric prior $\mathcal{L}_{\textrm{geom}}$ (\cref{eq:lossgeom}), pixel loss $\mathcal{L}_{\textrm{pix}}$ (\cref{eq:img_loss}), appearance prior $\mathcal{L}_{\textrm{in}}$ (\cref{eq:inpaint_loss}) with $\lambda_{(\cdot)}$ as weight terms:

\vspace{-0.3cm}
\begin{equation}
\mathcal{L}= \lambda_{\textrm{geom}} \cdot \mathcal{L}_{\textrm{geom}}
    + \lambda_{\textrm{in}} \cdot \mathcal{L}_{\textrm{in}} + \mathcal{L}_\textrm{pix}.
    \label{eq:overall_loss}
\end{equation}

\section{Experiments}
\label{sec:exp}
In this section, we evaluate \MethodName{} on various real-world captured datasets for text-guided inpainting.

\noindent\textbf{Datasets.}
 We take 360-degree datasets from MipNeRF, MipNeRF-360, and NeRFStudio~\cite{barron2021mip, barron2022mip, tancik2023nerfstudio}. 
 In addition, due to the absence of ground truth data for 360\degree~scene inpainting, we capture new datasets with and without the object removed for \textit{quantitative} evaluation. 
Segmentation mask ground truth does not exist for the 360\degree~scenes we evaluate. 
We also compare with front-facing datasets from SPIn-NeRF~\cite{spinnerf} and IBRNet~\cite{wang2021ibrnet} to show that our method produces better inpainting over baseline methods.

\noindent\textbf{Baseline.}
We select baselines based on specific tasks. SPIn-NeRF \textbf{SPN}~\cite{spinnerf} is the closest work to ours. For segmentation, we compare with the multiview segmentation from \textbf{SPN} and a recent video segmentation method \textbf{Dino}~\cite{zhang2022dino}. 
For the inpainting task, we compare our inpainting quality with our implemented version of SPN that works with 360-degree scenes \textbf{SPN-360}. We also compare with \textbf{per-frame image inpainting}~\cite{rombach2022high} to show that our method generates more consistent inpainting across different viewpoints.

\subsection{Segmentation Comparison}
We qualitatively compare our segmentation results to SPN and Dino. 
As input, we give the first frame segmentation to both SPN and Dino, and give the corresponding text instructions for \MethodName{}. As we are evaluating the extrapolation ability of each segmentation method over drastically different viewpoints, we select the 82nd frame from \textit{vasedeck} and the 81st image from \textit{room}.
They represent the respective frames of the videos created from the dataset.

\begin{figure}
  \centering
  \includegraphics[width=0.9\linewidth]{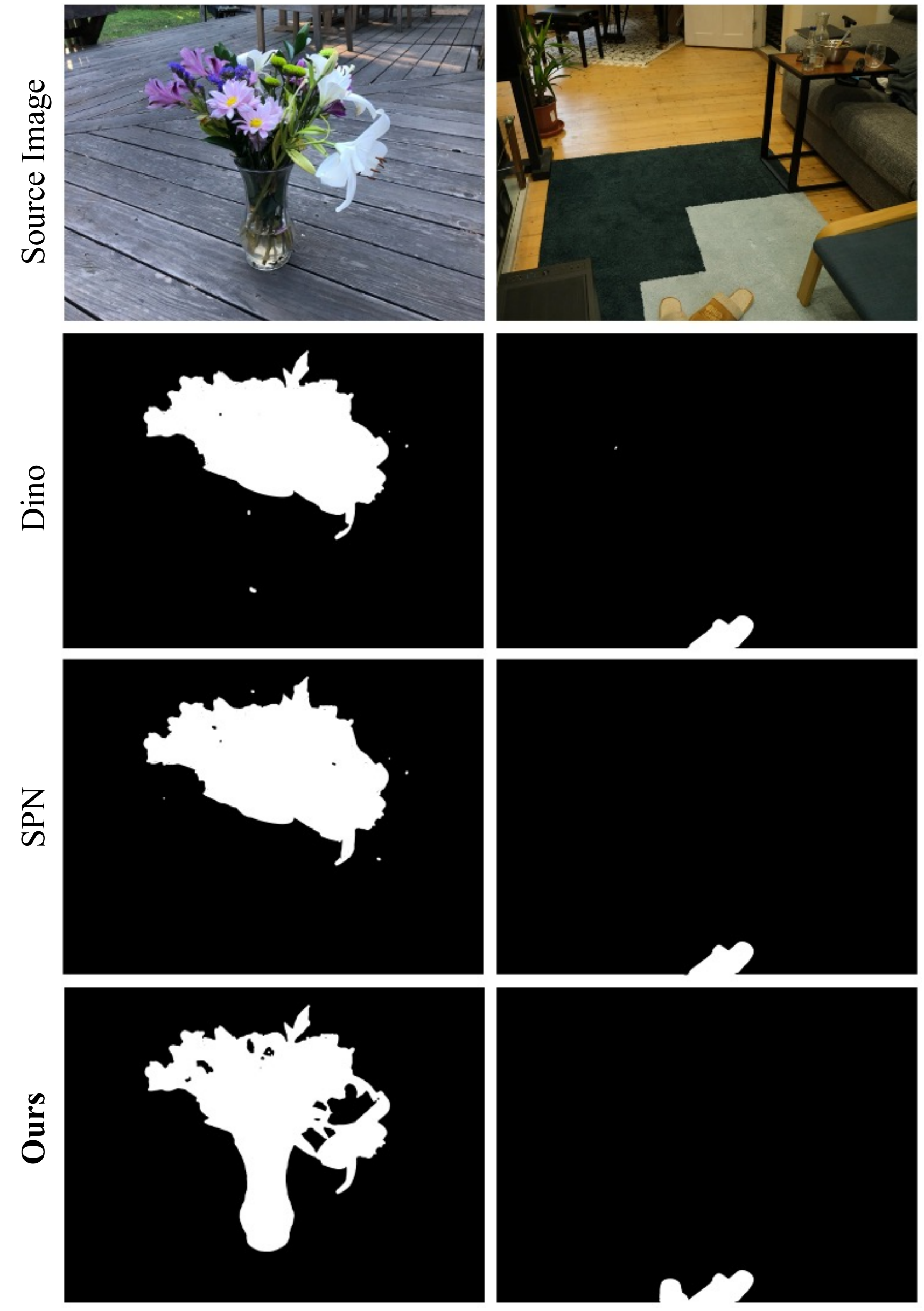}
  \vspace{-0.3cm}
  \caption{\textbf{Qualitative comparison on 3D object segmentation. } \MethodName{} outputs accurate masks for complex cases containing transparent (vase) or incomplete objects (partial slippers).}
  \label{fig:segmentation}
  \vspace{-0.4cm}
\end{figure}

As shown in \cref{fig:segmentation}, Dino produces incomplete segmentation for unseen views in challenging scenes, such as the \textit{vasedeck} featuring a transparent vase.
In such cases, SPN struggles to generate complete segmentation when initialized with Dino's output. 
While we also initialize with a 2D segmentation model (SAM), it facilitates flexible, promptable input. 
This can be utilized by our depth-warping refinement to output point prompts specifically in the vase section of the image. 
These prompts guide SAM towards accurate segmentation, as elaborated in \cref{sec:depth}. 
For \textit{room}, where only a part of the slippers is present in the image, Dino and SPN once again yield incomplete segmentations. \MethodName{} is able to output complete object segmentation. 

\subsection{Inpainting Evaluation}
\begin{figure*}
  \centering
  \includegraphics[width=0.9\linewidth]{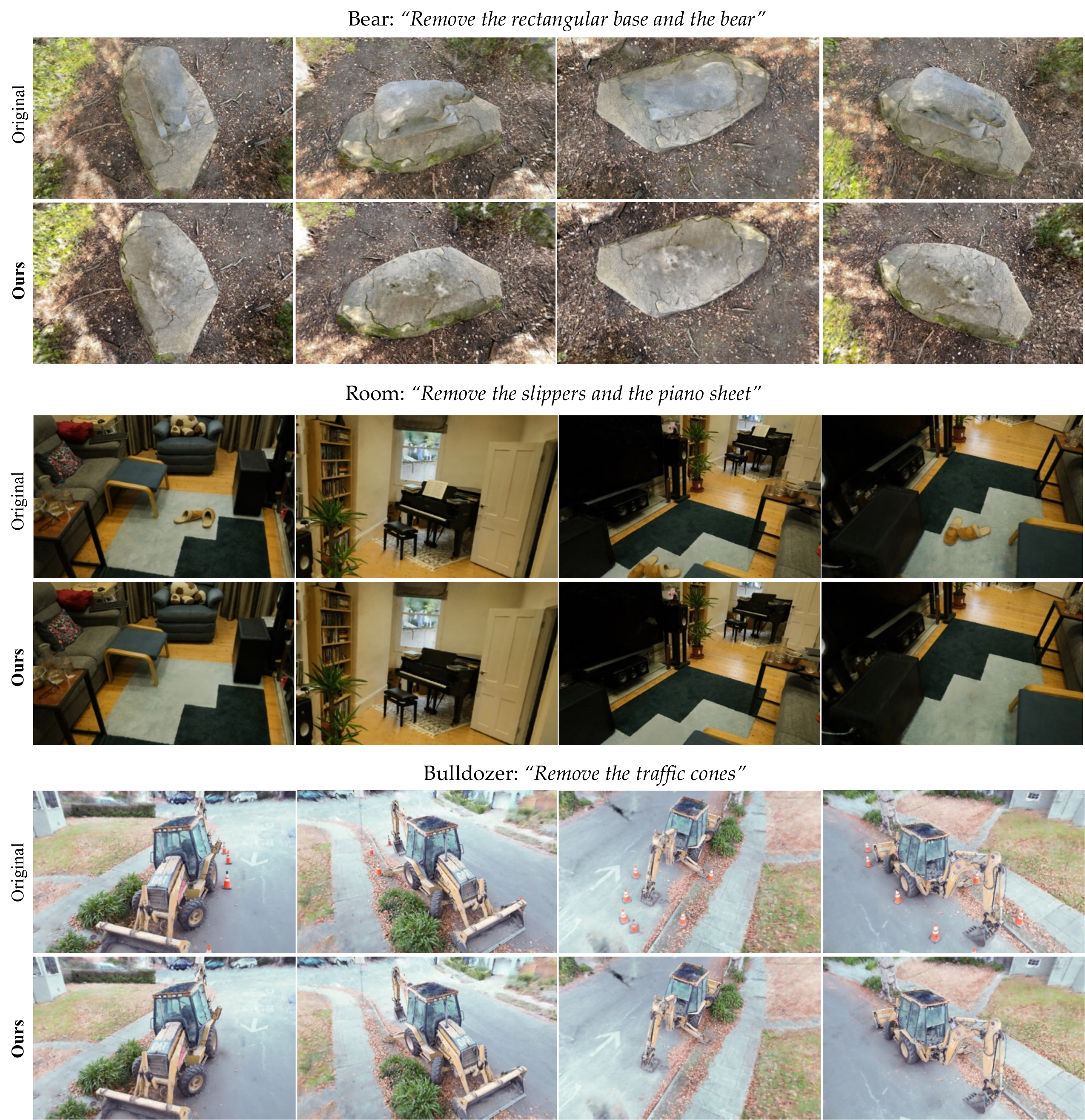}
  \caption{\textbf{Qualitative inpainting results on 360 scenes.} Our method works with various types of NeRF scenes. We can also remove arbitrary numbers of objects given the text input, independent of the complexity of the scene content.}
  \vspace{-0.3cm}
  \label{fig:res}
\end{figure*}

\noindent\textbf{Results on 360-degree scenes.}
\MethodName{} can handle a wide range of scenarios for object inpainting in 3D scenes, as depicted in~\cref{fig:res}. We encourage the readers to view supplementary videos to inspect the quality of our results.

\MethodName{} can remove \textit{large scene objects against complex backgrounds}. In \textit{Bear}, the background of the bear contains varying trees and branches, and the 2D inpainted images are therefore very noisy. However, with our floater removal and appearance prior, \MethodName{} produces a clean stone surface in the final edited NeRF.

\textit{Occlusion and geometry deformation} across varying viewpoints are major causes of 3D inconsistency in inpainting, and the datasets we use contain both scenarios. 
In \textit{Room}, the window, wall, and the piano behind the score can lead to inconsistency. However, we generate view-consistent content to seamlessly fill the missing region. 

\begin{figure}
  \centering
  \includegraphics[width=0.9\linewidth]{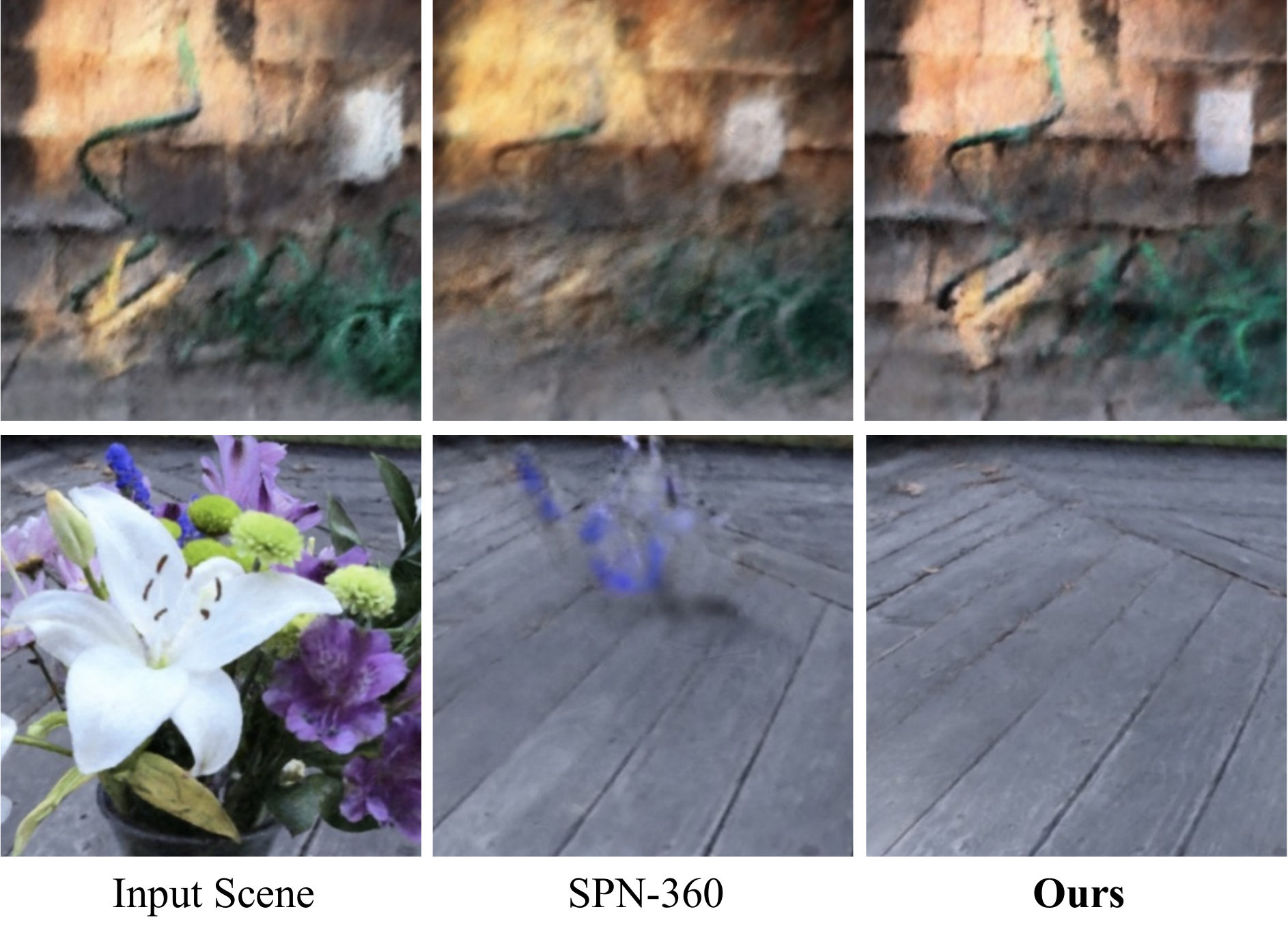}
 \vspace{-0.4cm}
  \caption{\textbf{Qualitative comparison with SPN-360.} Text: \textit{Remove the vase and the flowers. }\MethodName{} inpaints clean and visually plausible regions while better preserving surrounding scenes.}
  \label{fig:spn360}
  \vspace{-0.3cm}
\end{figure}

\begin{table}[]
\scalebox{0.85}{
    \centering
    \begin{tabular}{lllll} 
        \toprule & \multicolumn{2}{c}{ Cup } 
  & \multicolumn{2}{c}{Starbucks} \\
  \cmidrule(lr){2-3}\cmidrule(lr){4-5}
        
 Methods  &  LPIPS $\downarrow $ & FID $\downarrow$ & LPIPS $\downarrow $ & FID $\downarrow$\\
 \cmidrule(lr){2-5}
        
 Per-Frame  &  0.6149  & 201.70  & 0.5981  &  260.93 \\
 SPN-360    &  0.6421  & 252.34  & 0.6278  &  215.28 \\
  \midrule
 NeRFacto &  0.7328   &  271.56     &    0.6832   & 258.39 \\
$+\mathcal{L}_{\textrm{in}}$  &  0.7137    &  210.57     &    0.6658   & 223.82 \\
$+\mathcal{L}_{\textrm{geom}}$  &  0.6197    &  189.57     &    0.5795   & 166.45 \\
$+\mathcal{L}_{\textrm{in}} + \mathcal{L}_{\textrm{geom}}$ (\textbf{Ours}) &  \cellcolor{green!25}\textbf{0.5377}  & \cellcolor{green!25}\textbf{159.76} 
                    & \cellcolor{green!25}\textbf{0.4523}  &  \cellcolor{green!25}\textbf{153.46} \\

        \bottomrule
    \end{tabular}
    }
    \vspace{-0.3cm}
    \caption{\textbf{Quantitative evaluation on the inpainting quality}. Our method achieves better results than baseline methods and our ablated settings on captured datasets. }
    \label{tab:360}
    \vspace{-0.5cm}

\end{table}

\begin{figure*}
  \centering
  \begin{subfigure}{0.61\linewidth}
    \includegraphics[width=\linewidth]{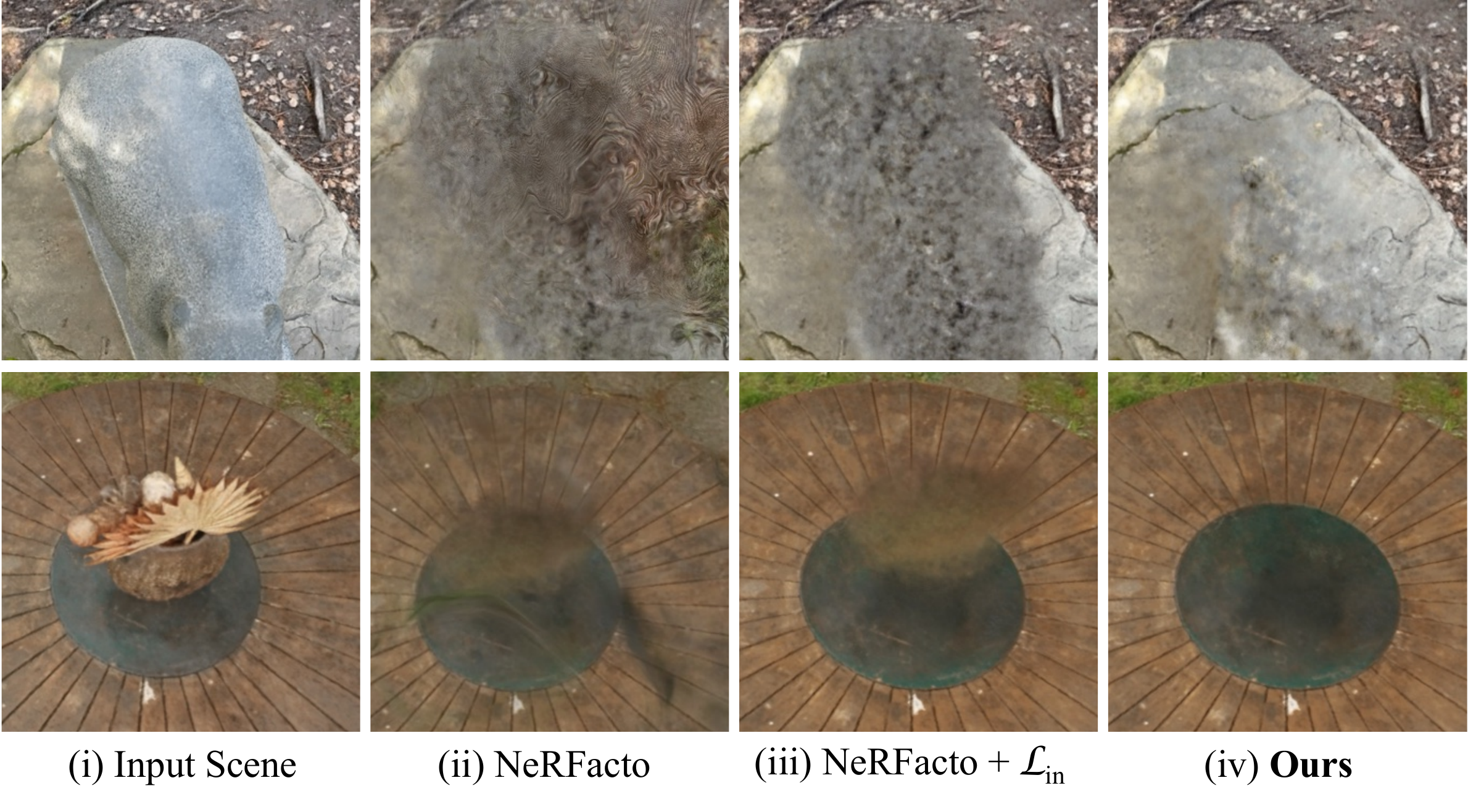}
    \vspace{-0.4cm}
    \caption{Qualitative ablation for $\mathcal{L}_{\textrm{geom}}$ on \textit{Bear} and \textit{Garden}.}
    \label{fig:ablation-fl}
  \end{subfigure}
  \hfill
  \begin{subfigure}{0.31\linewidth}
    \includegraphics[width=\linewidth]{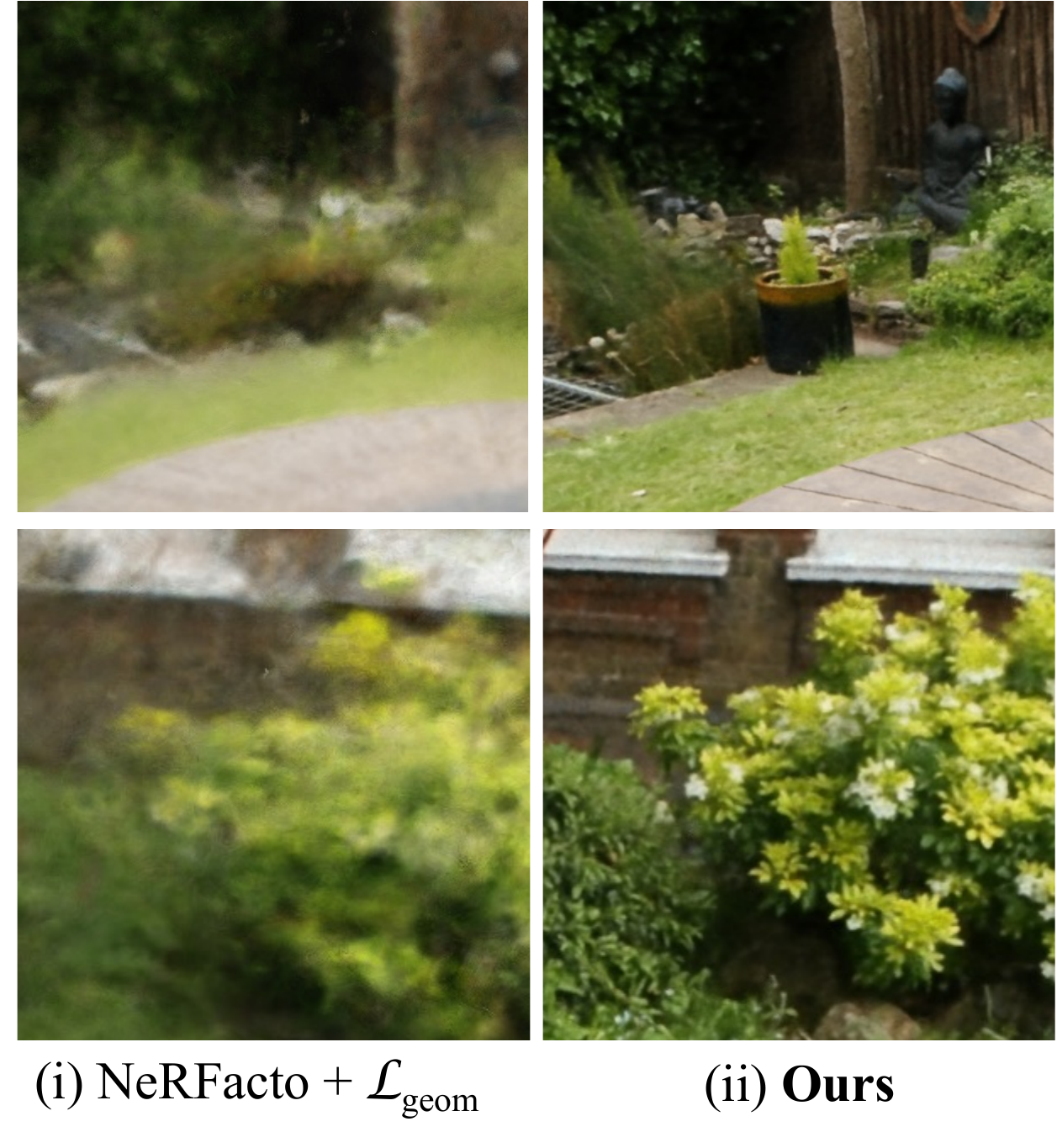}
    \vspace{-0.4cm}
    \caption{Qualitative ablation for $\mathcal{L}_{\textrm{in}}$ on \textit{Garden}. }
    \label{fig:ablation-lp}
  \end{subfigure}
  \vspace{-0.3cm}
  \caption{\textbf{Ablation for losses on geometric and appearance priors}. The artifact in the inpainted region is not as pronounced if viewed from aside as when viewed from the top. Our method is able to optimize an inpainted NeRF without artifacts and with a consistent and unblurry background. }
  \vspace{-0.5cm}
  \label{fig:ablation}
\end{figure*}

Moreover, \MethodName{} is capable of inpainting \textit{multiple objects} located \textit{anywhere in the 3D scenes} without introducing blurry artifacts in the resulting NeRF scene. When given a text input containing multiple objects (\textit{Room}: ``slippers'' and ``piano sheet''; \textit{Bulldozer}: multiple ``cones''), \MethodName{} produces inpainted regions that seamlessly blend with the surrounding context, yielding visually coherent and high-quality inpainting results. 

\cref{fig:spn360} shows qualitative comparison to \textbf{SPN-360}. Our method not only synthesizes a perceptually-consistent inpainted region, but also preserves the surrounding background closer to the input NeRF scene. We speculate the reason for the background-preserving inpainting to be that SPN inpaints on depth maps with segmentation masks generated from RGB images. We elaborate on our choice not to use 2D depth map inpainting in the supplementary.

\noindent\textbf{Ablation studies on our design choices. } 
Our depth-warping method produces refined point-based prompts for the segmentation model, and outputs complete and consistent multi-view segmentation, as shown in~\cref{fig:depth}. 

\cref{fig:ablation} qualitatively ablates our choice of loss functions. In \cref{fig:ablation-fl} (ii), the vanilla NeRFacto model outputs a concentrated artifact in the inpainted region along with noisy texture in nearby regions which we suspect is due to per-image appearance encoding on inconsistent 2D inpainted images. \cref{fig:ablation-fl} (iii) shows NeRFacto $+ \mathcal{L}_{\textrm{in}}$ which improves inpainted texture, but cannot reduce the floater artifact. These artifacts have view-dependent appearances from individual views and are therefore difficult to remove from appearance priors. In \cref{fig:ablation-fl} (iv) for \MethodName{}, we can see a clean and perceptually consistent surface in the edited scene. 
In \cref{fig:ablation-lp} shows \MethodName{} versus trained without $\mathcal{L}_{\textrm{in}}$. We can see that the LPIPS loss can improve blurry background output due to inconsistent 2D inpainting.
The lower part of \cref{tab:360} shows quantitative ablation on each loss term in \MethodName{}. Our complete architecture performs better than without each of the loss terms.  

\begin{table}
    \centering
    \scalebox{0.85}{
    \begin{tabular}{cccccc}
    \toprule
    Datasets  & Garden & Room & Vasedeck & Bulldozer & Bear \\
    \midrule
    \textbf{Ours} & 
    \cellcolor{green!25}\textbf{89\%} & \cellcolor{green!25}\textbf{71\%} & \cellcolor{green!25}\textbf{81\%} & \cellcolor{green!25}\textbf{83\%} & \cellcolor{green!25}\textbf{92\%}\\
    Per-frame & 11\% & 29\% & 19\% & 17\% & 8\%\\
    \bottomrule
    \end{tabular}
    }
    \vspace{-0.3cm}
    \caption{\textbf{User study comparing with per-frame inpainting on visual consistency between consecutive frames.} In each of the scenes, our inpainted NeRF renders higher view consistency than per-frame inpainting. Per-frame editing lacks a 3D understanding of each scene and inpaints each image independently. }
    \vspace{-0.6cm}
  \label{tab:clip}
\end{table}

\begin{figure}
  \centering
  \includegraphics[width=0.9\linewidth]{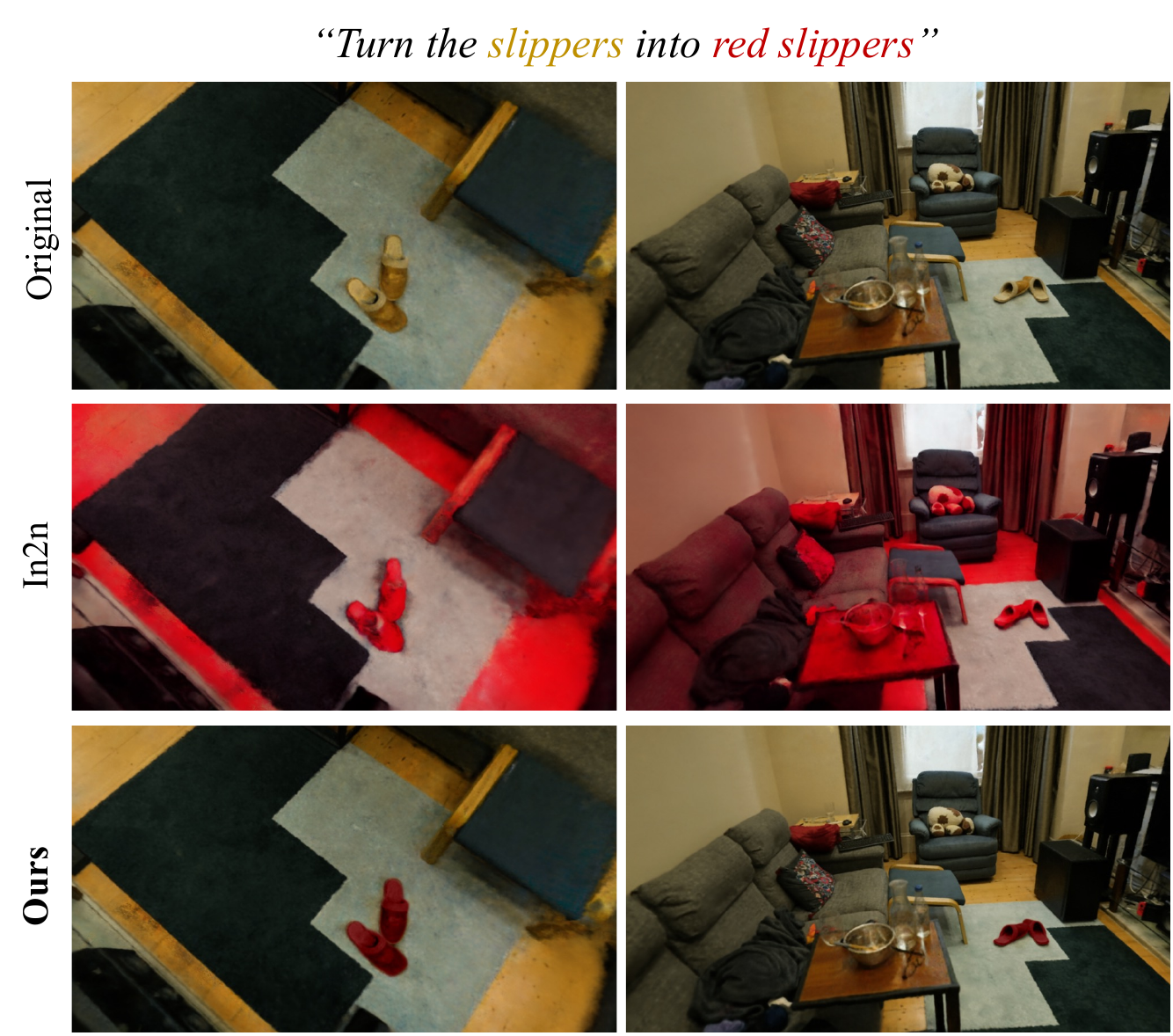}
  \vspace{-0.3cm}
  \caption{\textbf{Editing comparison with In2n.} \MethodName{}, combined with appropriate mask-conditioned image editing models, can generate accurate editing on desired objects. In contrast, In2n gives the wrong texture to unwanted regions.}
  \label{fig:in2n}
  \vspace{-0.5cm}
\end{figure}

\noindent\textbf{Inpainting quality.}
Due to the lack of baseline and ground truth datasets on inpainting 360\degree~NeRF scenes, we captured real-world datasets for quantitative comparison on the quality of inpainted renderings. Since \MethodName{} generates consistent and complete 3D segmentation over baseline methods, the 2D inpainting initialization is naturally much less noisy than baseline methods.

We evaluate our inpainting quality on each frame of the renderings with per-frame inpainting and SPN-360. 
We report LPIPS~\cite{zhang2018perceptual} and Frechet Inception Distance (FID)~\cite{heusel2017gans} metric in Tab.~\ref{tab:360} by comparing with the output of the captured empty scene rendered under the same camera trajectory which we use as ground truth. Our method outperforms each baseline method and outputs visually consistent inpainting without visual artifacts. 

A quantitative baseline comparison to SPN on frontal scenes is also in the supplementary materials. 

\noindent\textbf{3D consistency over per-frame inpainting.} A naive approach for 3D scene inpainting is to independently inpaint every rendered image of the scene with a 2D image inpainter. 
In contrast, \MethodName{} produces inpaintings with higher view consistency across all viewpoints.

We verify such a claim with a user study where participants were presented with two video clips of each inpainted scene, rendered with sequential camera trajectories. They were then asked to identify which clip appeared more consistent. Additional details about the user study can be found in the supplementary material. The results, presented in ~\cref{tab:clip}, clearly show that our rendered inpaintings exhibit superior temporal consistency compared to per-frame edits.

\vspace{-0.2cm}
\subsection{Editing Accuracy}
As shown by \cref{fig:in2n}, our segmentation module can be connected with a mask-conditioned image editor~\cite{couairon2022diffedit} to generate view-consistent editing with object-level control through text instructions, which InstructNeRF2NeRF (In2n)~\cite{haque2023instruct} cannot. However, note that editing is not the focus of our work. We show this result simply to demonstrate a possible extension to our method. Details are provided in the supplementary.

\vspace{-0.2cm}
\section{Limitations and Conclusion}
\vspace{-0.2cm}
\noindent\textbf{Limitations.} 
Our method inherits certain constraints of vision-language models. In scenarios where the text instruction cannot be accurately localized within the image, \MethodName{} may struggle in generating segmentation consistently aligned with the views. This issue arises when the initial masks provided by the 2D object detector are inaccurate or too noisy for effective refinement. Addressing this challenge is a focus of our future work.

\noindent\textbf{Conclusion.}
In conclusion, we have presented \MethodName{}, a unified system to accurately segment and inpaint objects in 360\degree~NeRFs with text instructions. We synthesize perceptually consistent inpainting without artifacts and can extend to object-level stylization, improving the controllability of NeRF.

\noindent \small{\textbf{Acknowledgement.} This work was supported in part by the Swiss National Science Foundation via the Sinergia grant CRSII5-180359.}

{
    \small
    \bibliographystyle{ieeenat_fullname}
    \bibliography{main}
}

\end{document}